\documentclass[letterpaper]{article}
\usepackage[letterpaper]{geometry}

\usepackage[utf8]{inputenc}
\usepackage[T1]{fontenc} 
\usepackage[cjk]{kotex}

\usepackage{amsmath}
\usepackage{graphicx}
\usepackage{url}
\usepackage{xcolor}
\usepackage{latexsym}

\usepackage{hyperref}
\definecolor{darkblue}{rgb}{0, 0, 0.5}
\hypersetup{colorlinks=true,citecolor=darkblue, linkcolor=darkblue, urlcolor=darkblue}
\usepackage{tabularx}
\usepackage{datetime}
\usepackage{arydshln}

\usepackage{setspace}
\usepackage{lineno}

\usepackage{booktabs}

\usepackage[authoryear,round]{natbib}
\usepackage{algorithmic,algorithm}
\renewcommand{\algorithmiccomment}[1]{\bgroup\hfill$\triangleright$~#1\egroup}

\usepackage[linguistics]{forest} 
\usepackage{synttree}
\usepackage{subcaption}
\usepackage{tikz-dependency}

\usepackage{langsci-gb4e}

\title{
Constituency Structure over Eojeol in Korean Treebanks
}
\author{ 
Jungyeul Park$^{1}$ and Chulwoo Park$^{2}$\\
$^{1}$KAIST, South Korea. {\tt jungyeul@kaist.ac.kr} \\
$^{2}$Anyang University, South Korea. {\tt cwpa@anyang.ac.kr}
}
\date{ }

\begin{document}

\maketitle

\doublespacing

\begin{abstract}
The design of Korean constituency treebanks raises a central representational question concerning the choice of terminal units. Although Korean words are morphologically complex, treating morphemes as constituency terminals can obscure the distinction between word-internal morphology and phrase-level syntactic structure, and can create mismatches with eojeol-based dependency resources. This paper argues for an eojeol-based constituency representation, with morphological segmentation and fine-grained part-of-speech information encoded in a separate, non-constituent layer. A comparative analysis shows that, under explicit normalization assumptions, the Sejong, Penn Korean, and KAIST treebanks can be compared over a shared eojeol-based constituency backbone. Building on this result, we outline an eojeol-based annotation scheme that preserves interpretable constituency, supports cross-treebank comparison and constituency–dependency alignment, and provides a surface-form terminal layer for future end-to-end Korean constituency parsing.
\end{abstract}

\tableofcontents

\section{Introduction}

A recurring issue in the design of Korean constituency treebanks concerns the choice of terminal units.
This choice is not merely a matter of annotation practice, but one that shapes how syntactic structure is interpreted, how constituency representations relate to one another, and how they interface with other syntactic resources.
In Korean, the question is particularly salient because surface words are internally complex, combining lexical material with multiple functional morphemes \citep{park-kim-2023-role}.
As a result, constituency annotation inevitably involves a decision about whether phrase structure should be built over morphemes or over larger grammatical units.

This paper examines the consequences of this decision and argues that defining constituency structure over eojeol\footnote{An \textit{eojeol} is the orthographic spacing unit in Korean, typically consisting of a lexical stem combined with one or more functional morphemes (such as case markers or verbal endings), which together form the basic unit of written word segmentation.} {offers a stable and interpretable terminal domain for Korean constituency treebank design}.
This stance is consistent with the synthetic view of Korean wordhood, according to which functional morphemes such as case particles and verbal endings do not constitute independent syntactic words but are grammatically integrated within a single syntactic unit \citep{chulwoo-2019-for-developing}.
{For the purposes of corpus annotation, eojeol can serve as surface-form units over which syntactic relations such as argument structure, modification, and clausal embedding are represented} \citep{park-kim-2024-word}.
Morphemes, while grammatically essential, contribute to syntax through their integration within eojeol rather than by entering syntactic configurations independently.
{If morphemes are used as constituency terminals,} word-internal morphological composition becomes intertwined with phrase-level organization, complicating the interpretation of hierarchical structure.

The goal of this paper is therefore to provide a theoretical and methodological foundation for constructing Korean constituency treebanks, clarifying which grammatical units should serve as syntactic terminals and how morphological information should be integrated without redefining constituency structure.
The claim advanced in this paper is representational rather than exclusionary.
We do not argue that Korean syntax is intrinsically eojeol-based, nor that functional morphology is syntactically inert.
Case particles, verbal endings, auxiliaries, and derivational morphemes encode information that is plainly relevant to syntactic interpretation, and several theoretical traditions treat such elements as syntactically visible.
The narrower claim is that syntactically relevant bound morphology need not be represented as independent constituency terminals in a treebank.
For the purposes of corpus design, cross-resource comparison, and constituency--dependency interoperability, eojeol terminals provide a stable syntactic backbone, while morphosyntactic information can be preserved in a separate annotation layer.\label{r2-1-1}

The motivation for an eojeol-based representation is reinforced by existing work on Korean dependency annotation \citep{noh-etal-2018-enhancing,seo-etal-2019-ud,han-etal-2020-annotation,park-etal-2021-klue,kim-etal-2024-kud}.
Prior studies have shown that dependency analysis relies on eojeol-based representations to maintain coherent syntactic relations and a stable interface with morphological information.
When constituency annotation adopts a different terminal granularity, mismatches arise between hierarchical and relational representations, making conversion, comparison, and evaluation across resources more complex than necessary.
The relation to Universal Dependencies (UD) should be understood in this limited sense.
The present proposal thus adopts an eojeol-based terminal layer in order to make Korean constituency annotation interoperable with UD-style resources.
Dependency annotation and constituency annotation retain different representational goals: the former encodes head-dependent relations, while the latter encodes hierarchical grouping.
The proposed format is therefore a constituency representation designed to share a stable token and category interface with eojeol-based dependency resources.\label{r1-1-1}

{A second motivation concerns constituency parsing.}
{One major purpose of a constituency treebank is to support parsers that assign phrase structure to input sentences.}
{For Korean, however, constituency parsing has often been evaluated under constrained conditions in which sentence boundaries and word or morpheme boundaries are assumed to match between system output and gold annotation.}
{Such a setting differs from end-to-end parsing, where the parser must operate from surface input and where segmentation decisions can affect downstream syntactic analysis.}
{An eojeol-based constituency representation therefore provides a surface-form terminal layer that is better suited to future end-to-end Korean constituency parsing, without requiring gold morpheme segmentation as a prerequisite for phrase-structure analysis.}

On this basis, we propose an eojeol-based constituency representation that aligns with eojeol-based dependency treebanks while retaining access to rich morphological information.
Under this approach, phrase structure is defined over eojeol terminals, while morphological segmentation and fine-grained POS information are recorded in a separate, explicitly non-constituent annotation layer.
This separation clarifies the linguistic interpretation of constituency, improves interoperability across Korean syntactic resources, and offers a principled basis for future corpus development without constraining morphological analysis.

\section{Background and conceptual distinctions}
\label{background-and-conceptual-distinctions}

This section lays out the conceptual background necessary for evaluating representational choices in Korean constituency annotation.
It clarifies how constituency and dependency representations differ in their treatment of terminal units and motivates why the choice of terminals is not a merely technical decision but a linguistically consequential one.
The section then situates eojeol and morphemes within Korean grammar, {identifying the grammatical considerations that motivate the representational stance adopted in this paper}.
Because several terms recur throughout the paper with distinct representational roles, Table~\ref{r2-minor-2} summarizes how they are used in the present discussion.

\begin{table}[!ht]
\centering
\footnotesize
\begin{tabular}{r p{0.7\textwidth}}
\toprule
\textit{Token} & A unit of corpus segmentation used for indexing and alignment, which may vary across annotation schemes \\
\textit{Surface form} & The overt string appearing in the sentence, recorded at the eojeol level in the proposed representation \\
\textit{Morpheme} & A lexical or functional unit inside an eojeol, represented in the morphological layer rather than as a constituency terminal \\
\textit{Terminal} & The leaf unit dominated by constituency structure, defined as an eojeol in the proposed scheme \\
\textit{Preterminal} & The category immediately dominating a terminal, represented by Universal POS in the proposed scheme \\
\textit{Syntactic atom} & The minimal unit over which phrase structure is built, identified with the eojeol in the proposed constituency representation \\
\bottomrule
\end{tabular}
\caption{Terminological distinctions}
\label{r2-minor-2}
\end{table}

\subsection{Constituency, dependency, and the choice of terminal units}

Constituency and dependency representations differ fundamentally in how syntactic structure is anchored to surface material.
In a constituency representation, terminal units define the atoms over which hierarchical phrase structure is built.
They determine what counts as the yield of a constituent, constrain how phrase-level grouping can be expressed, and shape the interpretation of nonterminal projections.
Changing the terminal unit therefore changes not only the leaves of the tree, but also the granularity at which syntactic generalizations are represented.

Dependency representations organize syntactic structure differently.
Rather than building phrase structure over terminals, they define binary relations between heads and dependents over a token sequence.
Frameworks such as Universal Dependencies therefore center syntactic analysis on labeled dependency arcs rather than on phrasal projections.
This difference is important for the present proposal: the aim is not to recast Universal Dependencies as a constituency grammar, but to define a constituency representation whose terminal and POS interface can remain compatible with UD-style annotation.\label{r1-1-2}

The choice of terminal units is consequently especially consequential for constituency treebanks.
Since terminals are the objects dominated by phrase structure, constituency treebanks that adopt different terminal units may differ in tree shape, constituent yield, and the granularity of syntactic encoding.
They are therefore not directly comparable without an explicit normalization of their terminal units, even when they annotate the same sentences.

This issue is particularly salient in Korean, where eojeol-based tokenization is standard in dependency treebanks and many downstream resources.
When constituency treebanks adopt morphemes as terminals \citep{chung-post-gildea:2010:SPMRL,choi-park-choi:2012:SP-SEM-MRL,park-hong-cha:2016:PACLIC,kim-park:2022}, while dependency treebanks operate over eojeol, conversion and evaluation require reconstructing or collapsing terminals, often through heuristic procedures \citep{xia-palmer-2001-converting}.
Such mismatches complicate cross-resource comparison and obscure whether observed differences reflect genuine syntactic variation or representational artifacts.

Comparable issues arise in the annotation of other morphologically rich languages, where the boundary between word-internal morphology and phrase-level syntax is not always straightforward.
In languages with rich inflection, cliticization, agglutination, or productive derivation, annotation projects must decide whether syntactic structure should be built over surface word forms, morphologically segmented units, or larger syntactic words.
These choices affect terminal granularity, the representation of functional morphology, and the comparability between constituency and dependency resources.
The Korean case examined in this paper is therefore not isolated, but it is particularly clear because eojeol provides an orthographically visible unit that can serve as a stable terminal backbone while preserving morpheme-level information in a separate layer.\label{r2-minor-3}

Distinguishing terminals as syntactic atoms from tokens as surface units is therefore essential.
A tokenization scheme may be suitable for corpus indexing, morphological analysis, or dependency annotation without necessarily being the appropriate choice for constituency terminals.
The present work adopts this distinction explicitly and treats the choice of constituency terminals as a representational decision grounded in syntactic considerations rather than in surface segmentation alone.

\subsection{Eojeol and morphemes in Korean grammar}

Eojeol occupies a central but complex position in Korean grammar.
Orthographically, eojeol corresponds to a spacing unit in written Korean.
Grammatically, it {can serve as the surface-form unit through which syntactic relations such as argument structure, modification, and case marking are represented in corpus annotation}.
At the same time, eojeol is internally complex, typically consisting of a lexical stem combined with one or more functional morphemes.

Because eojeol exhibits internally heterogeneous composition, it cannot be reduced to a uniform syntactic category and may function as a nominal, a verbal predicate, a modifier, or a clausal unit.
Its syntactic role is determined by the combination of lexical material and functional morphology, rather than by any single morpheme in isolation.
This makes eojeol {a useful terminal unit for representing surface-form constituency in Korean}, even though it is not morphologically atomic.

Morpheme segmentation is nevertheless indispensable for Korean linguistic analysis.
Morphemes encode case, tense, mood, derivation, and other grammatical information that is central to the interpretation of sentences.
They are essential for morphological analysis, generation, and many downstream tasks.
However, {in the proposed constituency architecture, morphemes are not treated as independent terminals; their syntactic contribution is represented through the eojeol in which they occur and through the separate morphosyntactic layer}.

Treating morphemes as constituency terminals therefore {risks conflating} two levels of grammatical organization.
It assigns syntactic status to units whose grammatical function is internal to words and whose distribution is governed by morphological rather than syntactic principles \citep{park-kim-2023-role}.
This {may obscure} the distinction between word-internal structure and phrase-level organization and {produce} constituency representations that are sensitive to segmentation choices rather than to syntactic relations.

For these reasons, eojeol {is adopted as the terminal unit in the proposed constituency representation}.
It aligns with the units used in Korean dependency resources, supports cross-resource comparison, and allows morphological information to be encoded explicitly without redefining the basic units of phrase structure.

\section{Existing Korean constituency treebanks}

Korean constituency treebanks differ from constituency resources for many other languages in that the relation between surface spacing units, morpheme sequences, and constituency terminals is not straightforward. In Korean, the eojeol provides the primary orthographic unit, and existing constituency treebanks generally refer to eojeol-level units at the terminal layer, either directly as terminals or indirectly through tokenized variants of the same surface unit. These terminals, however, are not simple word-level atoms. They usually encode an internal sequence of morphemes, each associated with a language-specific POS label. An eojeol terminal can therefore be understood as a compact representation of an expanded eojeol-internal subtree, whose children would be POS-labeled morpheme units.

The major Korean constituency treebanks differ in how this eojeol-internal information is exposed. In the Sejong and Penn Korean treebanks, the eojeol appears as the terminal unit, while the morpheme sequence and POS tags are encoded inside the terminal string. In the KAIST treebank, functional morphemes such as suffixes are explicitly separated from the lexical stem, so that an eojeol may no longer correspond to a single terminal node. Even in this case, however, the relevant unit remains eojeol-based in the sense that prefixes, roots, compounds, and functional morphemes are represented as parts of the same surface spacing unit. The sequence of internal morphemes may also represent a canonical morphological form that diverges from the surface orthographic realization.

This design contrasts with constituency treebanks for languages such as English \citep{marcus-etal-1993-building}, French \citep{abeille-clement-toussenel:2003}, or Chinese \citep{xue-etal-2005-ctb}, where terminal nodes generally correspond to surface words and their immediate parents are POS tags, with no comparable word-internal morphological structure visible in the constituency tree. Korean constituency trees therefore encode morphological composition at or near the terminal layer in a way that interacts with the representation of phrase structure, rather than relegating morphology entirely to a separate preprocessing step.

For the purpose of constituency parsing, it is common practice to convert Korean constituency treebanks into a representation where morphemes, rather than eojeol, serve as terminal nodes, each immediately dominated by a part-of-speech category, thereby aligning the tree shape more closely with cross-linguistic parsing conventions \citep{chung-post-gildea:2010:SPMRL,park-hong-cha:2016:PACLIC,kim-park:2022}. This conversion is useful for adapting Korean data to existing parsing architectures, but it also makes parsing dependent on a morpheme-level terminalization that is not directly available from surface input without morphological analysis.

In this section, we present the Korean constituency treebanks as they are originally defined, focusing on how eojeol and morphological information are integrated into phrase structure in each resource.

\subsection{The Sejong treebank}

The Sejong treebank adopts eojeol as the terminal unit at the constituency level. More precisely, each leaf corresponds to a single eojeol position in the surface sentence, but the terminal string is not a plain orthographic eojeol. It encodes the eojeol as an ordered sequence of morphemes with language-specific POS labels. The terminal can therefore be viewed as a compact representation of an eojeol-internal morphological subtree, while constituency structure itself is built only above the eojeol layer. This design choice is visible in the example, where inflectional endings, derivational suffixes, and punctuation are linearly attached to lexical material within the same terminal.

Two properties follow directly.
First, punctuation is not structurally independent {in the constituency tree}.
Sentence-final markers such as ./SF are incorporated into the same terminal as the verbal ending, rather than projected as separate syntactic units.
As a result, punctuation {does not independently determine} constituent boundaries or attachment decisions.
Second, phrase structure is obligatorily binary.
The Sejong treebank enforces {a binary branching annotation scheme}, even when the linguistic motivation for particular binarization decisions may not be transparent.
Complex nominal sequences are recursively decomposed into nested NP projections (e.g. NP $\rightarrow$ NP NP), yielding deeply right- or left-branching structures.
Because binarization is applied only above eojeol terminals, internal morphological relations are {not represented as part of the constituency structure}.

In sum, Sejong encodes morphology as {terminal-internal information}, while syntax is expressed through binary phrase structure over eojeol.
This yields trees that are uniform and compatible with parsing techniques that assume binary branching, such as CKY-style chart parsing \citep{cocke:1969,kasami:1966,younger:1967}, but it also means that morphosyntactic information is not projected as independent constituency structure.

\begin{figure}[!ht]
    \centering
\resizebox{\textwidth}{!}{
\footnotesize{
\begin{forest}
[S      [NP-SBJ         [NP     [NP-MOD [프랑스/NNP\\+의/JKG]]
[NP     [VNP-MOD [세계/NNG+적/XSN\\+이/VCP+ㄴ/ETM]]
[NP     [NP [의상/NNG]]
[NP [디자이너/NNG]]]]]
[NP-SBJ         [NP [엠마누엘/NNP]]
[NP-SBJ [웅가로/NNP\\+가/JKS]]]]
[VP     [NP-AJT         [NP     [NP     [NP [실내/NNG]]
[NP [장식/NNG\\+용/XSN]]]
[NP [직물/NNG]]]
[NP-AJT [디자이너/NNG\\+로/JKB]]]
[VP [나서/VV+었/EP\\+다/EF+${.}$/SF]]]]
\end{forest}
}
}
\caption{Constituency tree from the Sejong treebank}
\label{sejong-treebank-example}
\end{figure}
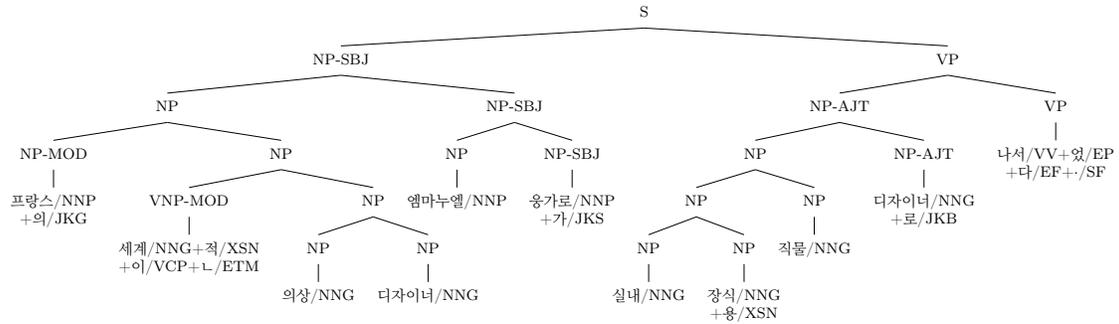

Figure~\ref{sejong-treebank-example} illustrates this representation for the sentence in \eqref{peulangseuui}, where eojeol function as {constituency terminals}, internal morphology {is encoded within each terminal}, and binary phrase structure is constructed above the terminal layer.\footnote{Glossing abbreviations follow standard practice: 
\textsc{gen} (genitive case; 의 \textit{ui}), 
\textsc{nom} (nominative case; 가/이 \textit{ga/i}), 
\textsc{purp} (purposive nominal derivation; 용 \textit{yong}), 
\textsc{pst} (past tense; 었- \textit{eoss-}), and 
\textsc{decl} (declarative sentence ending; 다 \textit{da}).}

\begin{exe}
\ex \label{peulangseuui}
\glll 프랑스의 세계적인 의상 디자이너 엠마누엘 웅가로가 실내 장식용 직물 디자이너로 나섰다. \\
\textit{peulangseuui} \textit{segyejeogin} \textit{uisang} \textit{dijaineo} \textit{emmanuel} \textit{unggaroga} \textit{silnae} \textit{jangsigyong} \textit{jikmul} \textit{dijaineoro} \textit{naseoeotda}. \\
France-\textsc{gen} world-\textsc{adj} costume designer Emmanuel Ungaro-\textsc{nom} interior decoration-\textsc{purp} textile designer-\textsc{cop} step.out-\textsc{pst}.\textsc{decl} \\
\trans `World-renowned French costume designer Emanuel Ungaro has stepped into the world of interior textile design.'
\end{exe}

\subsection{The Penn Korean treebank}

The Penn Korean treebank inherits the representational philosophy of the Penn treebank tradition and adapts it to Korean while preserving a range of crosslinguistic annotation conventions \citep{han-EtAl:2002}.

As in the Sejong treebank, each terminal corresponds to a single eojeol position in the constituency yield. The terminal string, however, is not a plain surface eojeol, but a morphologically analyzed representation containing a sequence of morphemes and language-specific POS labels. Although words are tokenized in a preprocessing sense, the terminals in the constituency structure correspond to eojeol-level units rather than to individual morphemes. Morphological information is therefore encoded inside terminal strings through compound POS annotations, rather than through independent syntactic projections.

The most distinctive characteristic of the Penn Korean treebank is its systematic use of null elements \citep{han-etal-2020-null}.
Empty categories such as \textit{pro} appear as explicit terminals within nominal or clausal constituents.
These null elements are not treated as peripheral annotations but as {structurally represented leaves}, occupying argument positions and encoding structural relations such as control and clausal complementation.
This practice aligns Korean annotation with Penn-style analyses developed for English \citep{marcus-etal-1993-building}, Chinese \citep{xue-etal-2005-ctb}, and other languages, and {supports crosslinguistic consistency within the Penn treebank tradition}.

A further characteristic concerns the treatment of functional morphemes.
Functional markers are consistently represented in a canonical form, regardless of phonological alternations conditioned by the preceding segment.
For example, the eojeol 정부는 \textit{jeongbuneun} (`government-\textsc{top}') is analyzed as 정부+은 \textit{jeongbu-eun} (`government-\textsc{top}') rather than 정부+는 \textit{jeongbu-neun} (`government-\textsc{top}'), abstracting away from surface allomorphy.
Other Korean constituency treebanks typically preserve such alternations in their annotations.
The Penn Korean treebank instead abstracts away from surface allomorphy in favor of morphologically normalized representations.

In addition, derivational morphology is explicitly encoded through part-of-speech distinctions.
Verbal derivations such as 발표+하+을 \textit{balpyo-ha-eul} (announce-\textsc{deriv}-\textsc{fut}) are annotated in a way that reflects category change, with the derivational element 하 \textit{ha} (`do') explicitly labeled as verbal.
This makes the transition from a nominal predicate to a verbal predicate explicit in the annotation, a distinction that is not systematically represented in other Korean constituency treebanks.

Unlike the Sejong treebank, the Penn Korean treebank does not strictly enforce binary branching.
While many structures are binarized, flatter configurations are permitted, particularly in argument and adjunct domains.
Nevertheless, the core assumption remains that syntactic structure is built over eojeol-level terminals, with internal morphology {encoded in terminal-level annotations rather than projected as independent constituency structure}.

In sum, the Penn Korean treebank can be characterized as structurally explicit and morphologically normalized.
It enriches phrase structure through null elements, canonicalized functional morphology, and explicit derivational labeling, while maintaining eojeol as the {terminal unit of constituency representation}.

\begin{figure}[!ht]
    \centering
\resizebox{\textwidth}{!}{
\footnotesize{
\begin{forest}
[S [NP-SBJ [영국/NPR\\+정부/NNC+은/PAU]] 
[VP [S-COMP [NP-SBJ [*pro*]] [VP [VP [NP-ADV [2/NNU] [일/NNX] [중/NNX]] 
[VP [NP-OBJ [NP [전/DAN] [칠레/NPR] [독재자/NNC] [아우구스토/NPR] [피노체트/NPR\\+의/PAN]] 
[NP [최종/NNC] [처리/NNC] [방침/NNC\\+을/PCA]]] 
[VV [발표/NNC\\+하/XSV+을/EAN]]]] 
[VX [것/NNX+이/CO\\+라/EFN+고/PAD]]]] 
[밝히/VV\\+었/EPF+다/EFN]] [${.}$/SFN]]
\end{forest}
}
}
\caption{Constituency tree from the Penn Korean treebank}
\label{penn-treebank-example}
\end{figure}
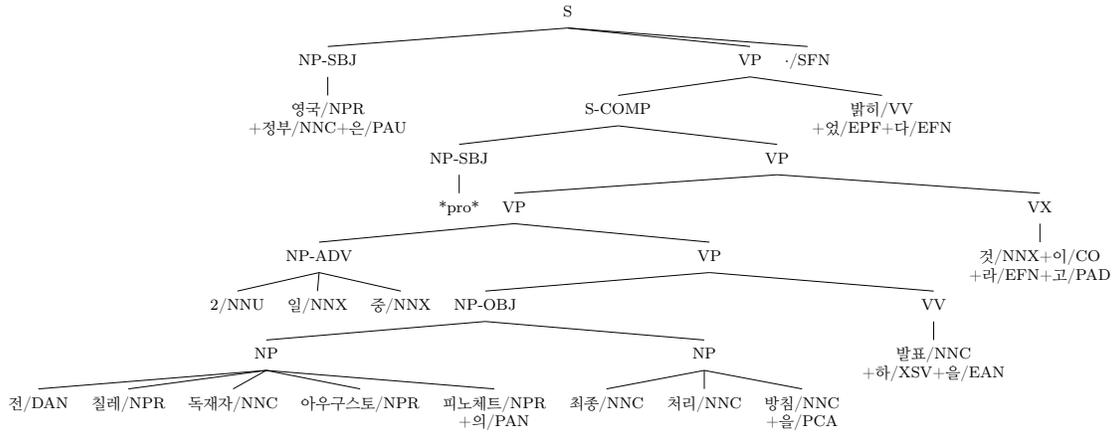

Figure~\ref{penn-treebank-example} illustrates this design for the sentence in \eqref{yeongguk-jeongbu}, where a null subject \textit{pro} is introduced in the embedded clause and clausal complementation is represented through an explicit S-COMP projection.

\begin{exe}
\ex \label{yeongguk-jeongbu}
\glll 영국정부는 2 일 중 전 칠레 독재자 아우구스토 피노체트의 최종 처리 방침을 발표할 것이라고 밝혔다. \\
\textit{yeongguk}\textit{jeongbuneun} 2 \textit{il} \textit{jung} \textit{jeon} \textit{chille} \textit{dokjaeja} \textit{auguseuto} \textit{pinocheteui} \textit{choejong} \textit{cheori} \textit{bangchimeul} \textit{balpyohal} \textit{geosirago} \textit{balkyeotda.} \\
UK.government-\textsc{top} two day during former Chile dictator Augusto Pinochet-\textsc{gen} final handling policy-\textsc{acc} announce-\textsc{fut} thing-\textsc{comp} state-\textsc{pst}.\textsc{decl} \\
\trans `The British government stated that on the 2nd it would announce its final policy on handling former Chilean dictator Augusto Pinochet.'
\end{exe}

\subsection{The KAIST treebank}

The KAIST treebank adopts a different design choice, in which an eojeol is not always represented as a single terminal \citep{choi-EtAl:1994}. Instead, functional morphemes may be separated from lexical roots and projected as independent terminals, so that word-internal morphological structure is partially exposed in the constituency tree. In this respect, KAIST can be understood as expanding some eojeol-internal morpheme/POS sequences into multiple terminal positions, rather than keeping them as compact terminal strings.

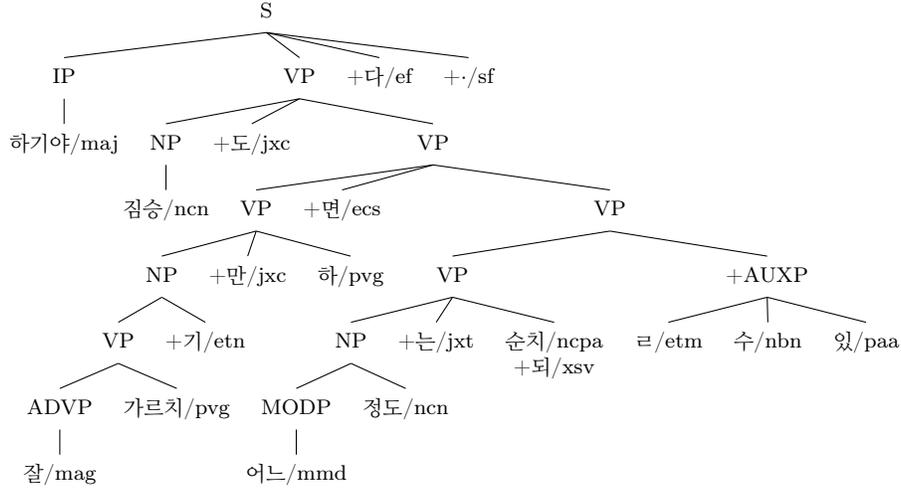
\begin{figure}[!ht]
    \centering
\footnotesize{
\begin{forest}
[S [IP [하기야/maj] ]
   [VP [NP [짐승/ncn] ][+도/jxc]
       [VP
           [VP
               [NP
                   [VP [ADVP [잘/mag] ] [가르치/pvg] ][+기/etn] ][+만/jxc]
               [하/pvg] ][+면/ecs]
           [VP
               [VP
                   [NP [MODP [어느/mmd] ] [정도/ncn] ][+는/jxt]
                   [순치/ncpa\\+되/xsv] ] [+AUXP [ㄹ/etm] [수/nbn] [있/paa] ]
                   ]]] [+다/ef] [+${.}$/sf] ]
\end{forest}
}
\caption{Constituency tree from the KAIST treebank}
\label{KAIST-treebank-example}
\end{figure}

This is immediately evident in the example in Figure~\ref{KAIST-treebank-example} for the sentence in \eqref{hagiyajimseungdo}. Suffixes such as case markers, auxiliary endings, and clause-linking morphemes appear as standalone nodes (도 \textit{do}, 만 \textit{man}, 면 \textit{myeon}, 다 \textit{da}), often attaching to larger projections rather than directly to lexical heads. When such functional material is projected separately, the eojeol no longer corresponds to a single terminal unit, but is distributed across multiple terminals within the tree.

\begin{exe}
\ex \label{hagiyajimseungdo}
\glll 하기야 짐승도 잘 가르치기만 하면 어느 정도는 순치될 수 있다. \\
\textit{hagiya} \textit{jimseungdo} \textit{jal} \textit{gareuchigiman} \textit{hamyeon} \textit{eoneu} \textit{jeongdoneun} \textit{sunchidoel} \textit{su} \textit{itda.} \\
well animal-\textsc{add} well teach-\textsc{nmlz}-\textsc{only} do-\textsc{cond} some degree-\textsc{top} tame-\textsc{pass} possibility exist-\textsc{decl} \\
\trans `Well, even an animal can be tamed to some extent if you just teach it well.'
\end{exe}

As a consequence, constituency and morphology are interleaved.
Verb phrases may dominate nominalized VPs, auxiliary constructions are explicitly layered, and clause boundaries are marked through functional projections rather than inferred from endings inside terminals.
The resulting trees represent derivational and functional morphology more explicitly than the Sejong or Penn Korean treebanks.

More broadly, this design choice is compatible with grammatical traditions in which functional morphemes are treated as syntactically visible elements rather than as purely morphological affixes.
By projecting case markers, auxiliaries, and clause-linking elements as independent nodes, the KAIST treebank represents aspects of word-internal structure within the constituency tree.
This alignment should be understood as representational rather than theory-testing, since the treebank remains a descriptive resource rather than an implementation of a specific grammatical model.

This choice also means that the eojeol is not always the terminal domain of constituency structure.
Phrase boundaries may be represented below the eojeol level, and syntactic heads may correspond to morphemes rather than surface-form units.
While this allows morphosyntactic information to be encoded directly in the tree, it complicates direct comparison with eojeol-based treebanks and requires parsers to model morphology and syntax within a single constituency representation.

In short, the KAIST treebank treats eojeol as a composite unit in many cases, and represents functional morphology as part of constituency structure.

\subsection{Comparative summary and discussion}

The three Korean constituency treebanks considered in this section adopt different assumptions about the relationship between morphology and phrase structure, leading to distinct formal properties at the level of nonterminal expansion.

The Sejong treebank enforces a strict Chomsky normal form, in which all nonterminal (NT) productions take the form NT $\rightarrow$ NT NT, and all nonterminals correspond to phrasal categories, {with} NT $\rightarrow$ terminal as a unary rule.
Terminals are eojeol-level units whose internal morphological composition is {encoded within the terminal rather than projected in the constituency tree}.
As a result, phrase structure is highly regular and uniformly binary, while morphosyntactic relations internal to eojeol are {represented outside the phrasal structure}.
This yields trees that are highly constrained and well suited to classical constituency parsing algorithms, {although morphology--syntax interactions are not encoded as independent constituency projections}.

The Penn Korean treebank occupies an intermediate position.
Its nonterminal expansions can be characterized schematically as NT $\rightarrow$ NT$^{+}$ terminal$^{\ast}$ or NT $\rightarrow$ terminal$^{+}$, with eojeol remaining the terminal unit.
Part-of-speech labels may also function as a preterminal layer mediating between nonterminals and terminals, rather than as independent syntactic categories.
Within this design, internal morphological distinctions are partially externalized through compound part of speech tags and explicit derivational relabeling, with terminals further decomposed into morpheme level units.
In particular, category-changing morphology is reflected by assigning derivational elements grammatical labels corresponding to their syntactic role, even though they are not projected as independent terminals.
In addition, the systematic use of null elements introduces further structural explicitness, allowing argument structure and clausal embedding to be represented in a way that is consistent with Penn-style analyses across languages; however, these null elements are conventionally removed during constituency parsing and are not treated as part of the derived surface structure \citep{bikel:2004:CL}.
At the same time, the Penn Korean treebank does not impose a strict binary branching constraint, permitting flatter structures in domains such as adjunct attachment.

The KAIST treebank adopts a different strategy.
Nonterminals may dominate both other nonterminals and terminal nodes, giving rise to expansion patterns of the form NT $\rightarrow$ (NT+ terminal$^\ast$)$^+$ | terminal$^+$.
Functional morphemes are projected as independent terminals, and eojeol is treated as a composite unit {distributed across multiple terminals in the tree}.
This allows derivational structure, auxiliary constructions, and clause-linking morphology to be represented {directly} within constituency structure.
While this {makes morphosyntactic information structurally explicit}, it reshapes branching assumptions and motivates parsing models that jointly model morphology and phrase structure.

These resources therefore exemplify three distinct strategies for balancing morphological information and constituency structure in Korean.
Sejong prioritizes formal regularity and eojeol-level constituency, Penn Korean emphasizes crosslinguistic structural consistency within the Penn treebank tradition while maintaining eojeol as the {terminal unit}, and KAIST foregrounds morphological transparency and derivational structure.
The resulting differences have direct consequences for treebank conversion, parser design, and cross-resource comparison.
In particular, assumptions about terminal granularity, branching constraints, and the syntactic visibility of morphology must be made explicit before results obtained on one treebank can be meaningfully compared or transferred to another.

\section{Eojeol and constituency in Korean} \label{eojeol-constituency-korean}

The choice of terminal unit is a foundational design decision in constituency annotation, because it determines how syntactic structure interfaces with morphology and how representations align across resources.
In Korean, this choice is nontrivial because eojeol simultaneously encodes lexical material and functional morphology, blurring the boundary between word-level syntax and internal morphological composition.
This section examines the implications of this tension for constituency representation.
It first argues that treating morphemes as constituency terminals introduces both conceptual and practical difficulties {for the resource-design goals pursued here}, and then articulates an eojeol-based representational stance in which constituency structure is defined over eojeol, while morphological information is preserved through explicit but non-constituent annotation layers.

\subsection{Why morpheme-based constituency terminals are problematic}

Treating morphemes as constituency terminals raises a representational issue for treebank design, but this should not be understood as a denial of morphology--syntax interaction.
Korean functional morphology, including case particles, verbal endings, auxiliaries, and derivational suffixes, is syntactically consequential and has been analyzed in several theoretical traditions as syntactically visible material.
The question is therefore not whether Korean functional morphology is relevant to syntax, but whether syntactically relevant bound morphology should be represented as independent constituency terminals in a treebank.
From the perspective of corpus design, projecting morphemes as terminals can blur the boundary between word-internal morphological composition and phrase-level syntactic grouping, especially when functional morphemes are assigned structural positions comparable to those of phrasal units.
\label{r2-1-2}\label{r2-4-1}

This decision has direct structural consequences.
When functional morphemes such as case particles, verbal endings, or derivational suffixes are projected as terminals, constituency trees may encode patterns of morphological attachment alongside phrase-level grouping.
The resulting structures can become deeper and more segmented, with unary or intermediate projections whose primary function is to host morphological material.
Such trees may introduce boundaries induced by word-internal composition in addition to phrase-level boundaries.
These effects may be appropriate under theories that explicitly project functional morphology, but they introduce additional representational commitments that are not required for an eojeol-based constituency backbone.

The problem is therefore not that morphology becomes visible, but that it is made visible {within} the constituency hierarchy.
Morphological distinctions internal to eojeol are then represented as constituency distinctions, inviting category assignments and hierarchical relations that may not have a stable phrase-structural interpretation {under the annotation architecture proposed here}.
Functional morphemes that do not independently project phrases are consequently placed in the same representational space as phrasal constituents, increasing terminal granularity {and making the interpretation of constituency boundaries less direct}.

Morpheme-based terminals also introduce systematic misalignment with existing Korean dependency resources, which are predominantly organized around eojeol-level tokens.
When constituency terminals do not correspond to eojeol units, aligning constituency trees with eojeol-based dependency structures requires reconstruction of word boundaries through merging, projection, or other conversion procedures.
Such procedures are often heuristic and not fully information preserving.
As a consequence, variation in morphological segmentation can be carried into syntactic evaluation, making cross-corpus comparison sensitive to preprocessing conventions rather than to genuine differences in syntactic structure.

This mismatch directly affects treebank conversion, cross-resource alignment, and comparative evaluation.
Metrics and analyses that presuppose a shared notion of terminal units no longer operate over a stable domain when constituency and dependency representations are defined over incompatible granularities.
Apparent syntactic discrepancies may therefore arise not from structural divergence, but from representational decisions concerning the exposure of morphemes.

The critique is thus not directed at morpheme-level annotation itself.
Morphological segmentation and detailed part-of-speech information are indispensable for Korean and should be retained.
The issue is specifically the use of morphemes as constituency terminals {when the goal is to define a surface-form constituency backbone aligned with eojeol-based dependency resources}.
This motivates a representational stance in which constituency is defined over eojeol, while morphological information is preserved in a separate, explicitly non-constituent layer.

\subsection{Toward an eojeol-based constituency representation with morphological annotation}

In response to the limitations identified above, this subsection outlines an eojeol-based approach to constituency representation that preserves a clear division of labor between syntax and morphology.
The proposal is intentionally representational rather than procedural, and is meant to clarify what constituency structure should encode in Korean, {as a resource-design choice}, independently of any particular annotation format or parsing algorithm.

Under this stance, eojeol is taken to be the basic terminal unit of constituency structure.
Phrase structure is defined over eojeol terminals, so that syntactic bracketing is represented over surface-form units rather than over internal morphological segments.
This choice maintains direct compatibility with eojeol-based dependency resources and allows constituency trees to be compared, aligned, and evaluated without requiring reconstruction of word boundaries or ad hoc merging operations.
Constituency labels and hierarchical relations are thus interpreted as phrase-structural objects, while word-internal structure is represented outside the constituency backbone.

At the same time, adopting eojeol as the terminal unit does not entail discarding morphological information.
Instead, morphological segmentation, fine-grained part-of-speech distinctions, and related features are recorded in a separate annotation layer that is explicitly non-constitutive for constituency.
This layer provides access to morpheme-level information without assigning each morpheme an independent terminal position in phrase structure.
Morphology is therefore preserved as linguistic information, but it does not determine constituency boundaries or projections {in the proposed representation}.

This separation allows syntactic and morphological analyses to interact without collapsing into a single representational space.
Constituency trees remain interpretable as models of phrase-structural organization, while morphological annotations can be consulted when needed for tasks such as generation, error analysis, or downstream semantic interpretation.
Because morphological segmentation and labeling are confined to a separate layer, they do not alter the shape of the constituency tree itself, reducing sensitivity to preprocessing conventions and improving cross-resource comparability.

{The conceptual motivation for this architecture can be summarized through the following five points.}

\paragraph{Representational objective of constituency}
Constituency annotation is intended to encode syntactic grouping and hierarchical relations among units that participate in phrase-level composition.
Its role is to model how phrases are formed and combined, not to represent internal morphological structure within words.
A constituency representation therefore benefits from a principled distinction between syntactic terminals and sublexical material whose contribution is mediated through those terminals.
When this distinction is not maintained, constituency structure may reflect patterns of morphological attachment rather than phrase-level organization.
Clarifying the representational objective of constituency is thus a prerequisite for determining an appropriate choice of terminal units.

\paragraph{Eojeol as the proposed syntactic terminal in Korean}
{In the proposed architecture, eojeol is adopted as the terminal unit for Korean constituency annotation.}
Syntactic dependencies, argument structure, and phrase-level grouping in existing Korean resources are commonly represented over eojeol-level units rather than over individual morphemes.
This makes eojeol a stable point of alignment between constituency structure and eojeol-based dependency representations.
Morphemes, by contrast, are represented as part of the internal morphosyntactic analysis of eojeol rather than as independent terminals in the constituency tree.
Treating eojeol as the terminal unit therefore {supports a surface-form constituency representation compatible with Korean dependency resources}, without imposing morpheme-level terminalization as a prerequisite for phrase-structure annotation.

\paragraph{Separation of constituency and morphology by design}
A central design choice of the proposed representation is the explicit separation between constituency structure and morphological annotation.
Morphological segmentation and fine-grained part-of-speech information are linguistically indispensable for Korean, but they need not be constitutive of constituency structure.
Encoding morphology in a parallel, non-constituent layer allows constituency trees to remain stable under alternative segmentation schemes or morphological analyses.
At the same time, morpheme-level information remains fully accessible when required, without redefining constituency terminals or projections.

\paragraph{Stability and cross-resource comparability}
Defining constituency over eojeol terminals provides a stable backbone for comparison across syntactic resources.
Because eojeol-based constituency aligns directly with existing eojeol-based dependency treebanks, comparison between representations does not require heuristic reconstruction of terminal units.
As a result, differences among constituency resources can be normalized at the level of annotation conventions rather than terminal granularity.
This improves comparability and interpretability in cross-corpus evaluation and analysis, helping distinguish representational choices from preprocessing artifacts.

\paragraph{Scope of the proposal}
The present discussion motivates a representational stance and specifies an annotation format, rather than advancing a comprehensive theory of Korean syntactic architecture.
The proposal is best understood as a corpus design and normalization strategy for Korean constituency treebanks.
It neither assumes that functional morphology is syntactically irrelevant nor excludes derivational, lexicalist, or morphosyntactic-interface analyses in which bound morphology is syntactically visible.
Its claim is instead that, for constituency treebank annotation, phrase structure can be defined over eojeol terminals while morphosyntactic information is retained in a parallel layer.
The aim is to clarify what the constituency backbone encodes, independently of how particular grammatical theories analyze the internal syntax or morphology of the eojeol.
\label{r2-1-3}

\section{An eojeol-based constituency annotation proposal}
\label{annotation-proposal}

This section presents a concrete annotation proposal that instantiates the eojeol-based representational stance motivated in the previous section.
The proposal should be understood as a principled annotation design, rather than as a claim about a single definitive standard for Korean constituency treebanks.
It specifies how constituency structure is constructed over eojeol terminals and how morphological information is recorded in a separate layer.
The goal is to define an annotation scheme that is explicit, interoperable across existing Korean resources, and suitable for corpus construction and evaluation, while preserving a linguistically interpretable notion of constituency.

\subsection{Constructing constituency over eojeol terminals}

The proposed representation defines constituency structure over eojeol terminals.
Each terminal corresponds to an overt eojeol in the sentence, and phrase structure is built over these terminals rather than over morpheme-level units.
Morphological segmentation, detailed part-of-speech information, and morphosyntactic features are retained, but they are represented in a parallel morphological layer rather than projected as independent constituency terminals.
In this sense, the proposal separates the constituency backbone from the internal morphological analysis of each eojeol.

This design has two consequences.
First, the yield of a constituency tree is an ordered sequence of eojeol terminals.
Second, the syntactic domain over which phrase structure is constructed is kept distinct from the morphological domain in which eojeol-internal composition is represented.
The proposed format therefore does not deny the syntactic relevance of Korean morphology.
Rather, it specifies that morphology is not used to define the terminal inventory of the constituency tree itself.

Existing Korean constituency treebanks can be related to this proposed representation through explicit normalization steps.
These steps should not be understood as the proposal itself, but as a way of showing how existing resources can be brought into the proposed eojeol-based constituency space.
The equivalence intended here is representational and conditional, not descriptive, formal, or empirical in a stronger sense.
It does not mean that the original treebanks encode the same grammatical analysis, nor that their annotation choices are theoretically interchangeable.
Rather, it means that, after explicitly specified normalization steps, their overt eojeol yields can be placed in a shared constituency space for the limited purposes of comparison, {alignment}, and annotation design.
The normalization therefore defines a common backbone over surface eojeol terminals, while leaving aside resource-specific commitments concerning null elements, binarization, and the syntactic projection of functional morphology.\label{r2-2-1}

We assume that Sejong and Penn Korean constituency treebanks can be mapped to this backbone under a set of explicit {normalization} transformations.
First, Sejong trees are debinarized.
Because Sejong enforces a strict Chomsky normal form, many phrase structures are realized through recursive binary expansions {introduced by the annotation format}.
Debinarization collapses these chains into flatter phrase structures without altering the overt eojeol yield.
This step removes structural depth introduced by a resource-specific binarization convention.

Second, Penn Korean trees are normalized by removing null elements.
Empty categories such as \textit{*pro*} are treated as annotation devices rather than as syntactic terminals in the proposed representation.
Removing null elements yields constituency trees defined exclusively over overt eojeol, while {preserving the phrase structure among the remaining overt constituents}.
This normalization aligns Penn Korean trees with eojeol-based resources that do not encode empty categories and avoids introducing abstract terminals that have no surface realization.

Third, eojeol tokenization is aligned across resources.
Both Sejong and Penn Korean treebanks ultimately operate over eojeol-level units, but differences in preprocessing conventions and token boundaries can obscure this commonality.
Under the proposed representation, constituency terminals are uniformly defined as surface eojeol, independent of internal morphological segmentation.
This alignment ensures that terminals correspond across resources and that constituency structure can be compared without reconstructing word boundaries.

Under this restricted interpretation, the normalization steps do not alter the overt eojeol sequence or its linear order, nor do they introduce new overt terminals.
They do, however, change the status of certain resource-specific structural analyses.
For this reason, the resulting equivalence should be understood as an equivalence of normalized surface-oriented constituency backbones, not as an equivalence of the original treebank analyses.
Sejong and Penn Korean can therefore be compared over a shared eojeol-based phrase-structural domain, while their original treatments of binarization, empty categories, and morphosyntactic detail remain distinct.\label{r2-2-2}

These transformations are not analytically neutral.
They preserve the overt eojeol sequence, the relative order of terminals, major phrase labels, and phrase-level dominance relations among overt constituents.
At the same time, they abstract away from information that is important in the original resources.
Debinarization removes intermediate binary nodes introduced by Sejong-style Chomsky-normal-form annotation, and may therefore suppress distinctions about local attachment if these distinctions are encoded only through binary branching.
Null-element removal in Penn Korean eliminates overt tree positions for unrealized arguments, traces, and empty categories, and therefore does not preserve the full analysis of argument realization, control, or long-distance dependency.
Collapsing morpheme-level terminals in KAIST reduces the visibility of functional morphology as independent syntactic material.
The proposed normalization is therefore not lossless.
Its justification is that the information discarded is not part of the specific representational layer targeted here, namely an eojeol-based constituency backbone over overt surface units.\label{r2-2-3}

A comparable normalization strategy has also been applied to the KAIST treebank.
Although KAIST exposes functional morphemes as independent terminals and allows eojeol-internal structure to enter the constituency tree, previous work has shown that these representations can be systematically converted into an eojeol-based format.
In particular, the SPMRL 2013 Shared Task describes a conversion procedure that reattaches functional morphemes, separated in the original KAIST annotation, to their host lexical items, yielding eojeol-level terminals and a correspondingly flattened constituency structure \citep{seddah-EtAl:2013:SPMRL}.
Under this transformation, KAIST trees are brought into a representational form that closely resembles debinarized Sejong-style constituency, with morphology encoded internally to terminals rather than projected syntactically.
{This conversion does not preserve all morphosyntactic distinctions encoded in the original KAIST trees, but it shows that KAIST can be mapped to an eojeol-based constituency backbone for the purposes of comparison and dependency alignment.}

\subsection{Terminal annotation with UPOS-labeled eojeol}

Under the proposed scheme, eojeol are the {only} terminals of constituency structure.
Each terminal is annotated with a universal part-of-speech label (UPOS) \citep{petrov-das-mcdonald:2012:LREC}, yielding terminals of the form \texttt{[UPOS [eojeol]]}.
UPOS labels function as preterminal categories: they mediate between phrase-level constituency labels and surface eojeol, while remaining external to the internal morphological composition of the eojeol itself.
This design replaces morpheme-level terminals without collapsing syntactic and morphological information, thereby preserving both syntactic interpretability and cross-resource compatibility.

The choice of UPOS as the preterminal category is motivated by both linguistic and practical considerations.
From a linguistic perspective, UPOS categories provide broad syntactic classes {for classifying the distributional behavior of eojeol in phrase structure}, such as nouns, verbs, adjectives, and adpositions.
These categories abstract away from language-specific morphological detail and characterize how eojeol function in syntactic configurations, independently of their internal segmentation.
Annotating eojeol with UPOS therefore aligns preterminal categories with syntactic behavior rather than with morphological realization.

From a resource and interoperability perspective, UPOS provides a stable and widely adopted inventory that facilitates comparison across corpora and formalisms.
Because UPOS labels are shared across many treebanks and annotation schemes, especially in Universal Dependencies \citep{de-marneffe-etal-2021-universal}, their use at the preterminal level allows eojeol-based constituency trees to interface naturally with dependency resources, evaluation tools, and downstream applications that already rely on these labels.
At the same time, restricting UPOS to the preterminal layer avoids overloading constituency structure with fine-grained distinctions that are better handled elsewhere.

This use of UPOS does not mean that UD dependency relations are imported into the constituency representation.
UPOS labels provide a coarse syntactic classification of eojeol terminals, while constituency labels such as \texttt{NP}, \texttt{VP}, and \texttt{S} encode phrase-level organization.
UD relations such as \texttt{nsubj}, \texttt{obj}, or \texttt{advmod} are not represented as constituency labels in the proposed scheme.
They may be aligned with the constituency representation through shared eojeol indices, but they remain part of a distinct dependency layer.
This design allows constituency and dependency annotations to be compared and {aligned} more transparently without collapsing their representational differences.\label{r1-1-3}

UPOS annotation is therefore complementary to, rather than a replacement for, detailed morphological information.
Fine-grained part-of-speech labels, inflectional features, and derivational structure are recorded separately in a non-constituent morphological layer.
This separation ensures that constituency terminals are classified according to their syntactic role, while language-specific morphological distinctions are preserved without influencing constituency bracketing or projection.

For example, a subject noun phrase headed by the eojeol 영국정부는 \textit{yeonggukjeongbuneun} (`the British government (\textsc{topic})') is annotated as \texttt{[NP-SBJ [NOUN 영국정부는]]}.\footnote{In the proposed scheme, composite institutional expressions such as 영국정부는 are annotated as \textsc{NOUN}, reflecting their syntactic behavior as nominal heads rather than their referential specificity.}
Here, the UPOS label \texttt{NOUN} reflects the syntactic category of the eojeol as a nominal argument, while the internal morphological composition of 영국정부는 is captured independently in the morphological annotation layer.
In this way, UPOS-labeled eojeol serve as syntactically meaningful terminals that anchor constituency structure while remaining compatible with detailed morphological analysis.

The use of UPOS is adopted here as a practical and cross-linguistically interpretable convention, not as a claim about the theoretical inventory of Korean lexical categories.
It also involves a deliberate trade-off.
Because UPOS provides a coarse and cross-linguistically oriented category inventory, it abstracts away from some language-specific distinctions encoded in Korean morphological tagsets.
For example, distinctions among lexical nouns, proper nouns, bound nouns, auxiliary predicates, case particles, verbal endings, and derivational suffixes cannot be fully represented at the UPOS preterminal level without either multiplying preterminal categories or reintroducing morpheme-level structure into the constituency tree.
In the proposed scheme, this loss is intentional but localized: UPOS classifies the syntactic behavior of the eojeol as a whole, while finer-grained Korean-specific distinctions are preserved in the XPOS-based morphological layer.
The proposal therefore prioritizes a stable and interoperable preterminal inventory for phrase structure, while retaining language-specific morphological categories outside the constituency backbone.\label{r2-5-1}

\subsection{Tokenization and internal nominal structure}

The proposed eojeol-based constituency representation requires an explicit definition of the surface units that serve as terminals.
For tokenization, the proposal adopts the surface-oriented guidelines articulated in \citep{park-tyers:2019:LAW}, which treat eojeol-level units as stable terminals while providing systematic conventions for punctuation, numerals, and other complex surface expressions.
Earlier resources exhibit substantial heterogeneity in this regard, ranging from symbol-level tokenization to ad hoc grouping of surface strings, often without explicit documentation of the underlying principles \citep{choi-park-choi:2012:SP-SEM-MRL}.
By committing to an explicit eojeol-based tokenization standard, the present scheme reduces annotation variance and improves cross-resource comparability without introducing task-specific preprocessing assumptions.

Beyond surface segmentation, the proposal also revisits the internal annotation of noun phrases.
Existing Korean constituency treebanks largely follow Penn Treebank-style flat noun phrase structures, in which adjectival modifiers and nominal elements are grouped without an explicit representation of modifier hierarchy.
Such flat analyses {provide a simple and consistent treatment of nominal expressions, but they leave modifier-internal structure underspecified}.
The present proposal therefore allows noun phrases to optionally encode nominal modifier structure through dedicated annotations such as \texttt{NML}, following refinements introduced in other treebank traditions \citep{vadas-curran-2011-parsing}.
These annotations are treated as part of the constituency representation itself rather than as post hoc enrichments, {when annotation guidelines support such finer nominal structure}.

The use of \texttt{NML}, however, is a design choice rather than a mandatory standard.
Its advantage is that it makes internal nominal modification more explicit than flat \texttt{NP} annotation, especially in complex nominal expressions where multiple modifiers precede a nominal head.
This can improve interpretability and may support more transparent comparison with dependency analyses of modifier attachment.
Its cost is that annotators must decide where modifier structure should be introduced, and different analyses may be possible for the same surface sequence.
A flatter \texttt{NP} analysis minimizes such decisions and may improve annotation consistency, but it leaves modifier-internal structure underspecified.
An \texttt{NML}-based analysis, by contrast, increases structural explicitness, but introduces additional attachment decisions that require clear guidelines and empirical validation.
For this reason, the present proposal treats \texttt{NML} as an optional refinement within the eojeol-based constituency backbone, not as a required component of the normalization procedure.\label{r2-5-2}

\subsection{Morphological annotation as a parallel, non-constituent layer}

Morphological segmentation and fine-grained part-of-speech information are recorded in a parallel annotation layer that is explicitly non-constitutive for constituency structure.
This layer captures word-internal structure and morphosyntactic detail without assigning such information a structural role in phrase-level representation.
Constituency trees are therefore defined independently of morphological segmentation, and remain stable with respect to how eojeol are internally analyzed.

In the morphological layer, each eojeol is segmented into its component morphemes and annotated using detailed language-specific part-of-speech labels (XPOS), following established conventions in Korean linguistic resources.
Segmentation is represented linearly, using a concatenative format such as \texttt{프랑스/NNP+의/JKG}, which makes explicit both lexical morphemes and functional elements.
This representation preserves information about case marking, derivation, inflection, and other morphological processes that are essential for linguistic analysis and downstream tasks.

Morphological units in this layer are aligned to eojeol terminals by index or identifier, but they do not introduce additional nodes, projections, or hierarchical structure into the constituency tree.
As a result, differences in morphological analysis, segmentation granularity, or tag inventories do not affect constituency bracketing or phrase structure.

This separation reflects a principled division of labor between syntax and morphology at the level of annotation.
Constituency structure encodes syntactic grouping and hierarchical relations among eojeol, while the morphological layer encodes internal word structure and fine-grained grammatical distinctions.
By keeping these layers formally distinct, the proposed scheme preserves rich morphological information while avoiding the representational complications that arise when morphemes are treated as constituency terminals.

This division of labor is not a claim that morphology and syntax are independent components of Korean grammar.
Nor is the morphological layer intended to make functional morphology peripheral or syntactically invisible.
Rather, it provides a separate locus for information that may be crucial for syntactic interpretation, including case marking, clausal endings, auxiliary constructions, derivational morphology, and morphosyntactic features.
The proposed architecture therefore allows morphology-sensitive analyses to be represented and queried, while preventing bound morphemes from automatically determining the terminal inventory and branching structure of the constituency tree.\label{r2-4-2}

\subsection{Integrated representation of constituency and morphology}

This subsection specifies how constituency structure and morphological information are represented jointly under the proposed eojeol-based annotation scheme.
The central design goal is to make both layers available in a single aligned representation, while keeping their formal roles distinct.
Constituency is encoded as hierarchical bracketing over eojeol terminals, whereas morphological segmentation and fine-grained part-of-speech information are recorded as aligned annotations that do not introduce constituency nodes.

Figure~\ref{annotation-example} illustrates the proposed annotation format.
The representation is organized into six aligned columns, each corresponding to a distinct annotation dimension.
Each row corresponds to a single surface eojeol or punctuation symbol, and all columns are indexed by a shared token identifier.
This alignment allows the constituency and morphological layers to be interpreted independently, while preserving a deterministic mapping between them.

The tabular representation is not intended as a new formalism for phrase structure, but as a CoNLL-like linearization of an ordinary constituency tree.
Comparable column-based encodings have been used in sequence-labeling approaches to semantic role labeling and other structured prediction tasks, where hierarchical or relational information is represented through token-aligned fields rather than through nested tree notation.
In the present proposal, the fifth and sixth columns jointly encode the same information as a bracketed constituency tree.
The left bracketing column specifies the sequence of opening brackets and node labels introduced before the current terminal, while the right bracketing column specifies the sequence of closing brackets after that terminal.
Given a well-formed bracket sequence, together with the token order in the \textsc{ID} column and the terminal labels supplied by the \textsc{UPOS} column, the full constituency tree is recoverable without ambiguity.\label{r1-2-1}

\begin{figure}[!ht]
    \centering
\footnotesize
\resizebox{\textwidth}{!}{
\texttt{
\begin{tabular}{llllll}
\multicolumn{6}{l}{\# sent id = BGAA0001-10012} \\
\multicolumn{6}{l}{\# text = 프랑스의 세계적인 의상 디자이너 엠마누엘 웅가로가 실내 장식용 직물 디자이너로 나섰다.} \\
BGAA0001-10012 & 프랑스의 & 프랑스/NNP+의/JKG & PROPN & 
(S (NP-SBJ (NML (AdjP (PROPN & ) \\
BGAA0001-10013 & 세계적인 & 세계/NNG+적/XSN+이/VCP+ㄴ/ETM & ADJ & (ADJ & )) \\
BGAA0001-10014 & 의상 & 의상/NNG & NOUN &  (NML (NOUN  & ) \\
BGAA0001-10015 & 디자이너 & 디자이너/NNG & NOUN & (NOUN & ))) \\
BGAA0001-10016 & 엠마누엘 & 엠마누엘/NNP & PROPN & (NP (PROPN & ) \\
BGAA0001-10017 & 웅가로가 & 웅가로/NNP+가/JKS & PROPN & (PROPN & ))) \\
BGAA0001-10018 & 실내 & 실내/NNG & NOUN &   (VP (NP-AJT (NML (NML (NOUN  & ) \\
BGAA0001-10019 & 장식용 & 장식/NNG+용/XSN & NOUN & (NOUN & )) \\
BGAA0001-10020 & 직물 & 직물/NNG & NOUN & (NOUN & )) \\
BGAA0001-10021 & 디자이너로 & 디자이너/NNG+로/JKB & NOUN & (NOUN & )) \\
BGAA0001-10022 & 나섰다 & 나서/VV+었/EP+다/EF & VERB & (VERB & )) \\
BGAA0001-10023 & . & ./SF & PUNCT & (PUNCT & )) \\
\end{tabular}}
}
\caption{Six-column joint annotation format for eojeol-based constituency and morphological analysis}
\label{annotation-example}
\end{figure}

The first column (\textsc{ID}) provides a unique token identifier, preserving sentence-internal order and enabling alignment with other annotation layers or external resources.
The second column (\textsc{Surface form}) records the surface eojeol form as it appears in the sentence.
Under the proposed scheme, surface forms are the {terminals} of constituency structure.

The third column (\textsc{Morphological segmentation (XPOS)}) encodes morphological segmentation and fine-grained, language-specific part-of-speech information.
Each surface form is decomposed into its constituent morphemes using a concatenative notation, with morphemes separated by plus signs and annotated with detailed XPOS tags (e.g., \texttt{프랑스/NNP+의/JKG}).
The XPOS inventory follows the Sejong POS tagset, which has become the de facto standard for Korean morphological annotation and is widely adopted across major Korean linguistic resources, including large-scale dependency treebanks used in both academic and industrial settings \citep{mcdonald-etal-2013-universal,park-tyers:2019:LAW}.
Adopting the Sejong tagset therefore ensures compatibility with existing Korean corpora and tools, while providing sufficient granularity to represent inflectional, derivational, and functional morphology.
In this column, morphological information is recorded as aligned annotation rather than as structural material, and therefore does not introduce nodes, projections, or hierarchical structure into the constituency tree.

The fourth column (\textsc{UPOS}) assigns a universal part-of-speech label to each surface form.
UPOS labels function as preterminal categories in the constituency representation, mediating between phrase-level categories and eojeol terminals.
These labels reflect the syntactic role of the eojeol as a whole, abstracting away from internal morphological composition.

The fifth and sixth columns encode constituency structure using a Penn-style bracketing convention.
The \textsc{LHS constituency} column records the opening of constituents at the current surface form position, including both phrase-level categories (e.g., \texttt{S}, \texttt{NP-SBJ}, \texttt{VP}) and preterminal UPOS nodes.
The \textsc{RHS constituency} column records the corresponding closing parentheses, indicating where constituents terminate.
Constituency boundaries are thus represented explicitly while remaining aligned with eojeol terminals.
Recoverability depends only on ordinary bracket well-formedness: every opening bracket must have a matching closing bracket, brackets must be properly nested, and each terminal must be associated with exactly one preterminal category.
If these conditions are violated, the representation is ill formed in the same sense that a Penn-style bracketed tree with unmatched or crossing brackets is ill formed.
Such cases are therefore annotation errors rather than genuine ambiguities of the format.\label{r1-2-2}

Figure~\ref{annotation-example} also illustrates how different annotation layers support distinct analytical tasks.
Constituency parsing operates over the surface eojeol sequence together with phrase-structural bracketing, and therefore requires only the \textsc{Surface form} column (Column~2), the \textsc{UPOS} column (Column~4), and the two constituency columns encoding left and right bracketing (Columns~5 and~6).
Under the proposed scheme, constituency structure is recoverable from these columns independently of morphological segmentation.

Morphological analysis, by contrast, is defined over the alignment between surface forms and their internal structure.
It therefore relies on the \textsc{Surface form} column (Column~2) together with the \textsc{Morphological segmentation (XPOS)} column (Column~3), which jointly encode eojeol-level tokens and their morpheme-level decomposition.
This corresponds to the format adopted in the morphologically analyzed Sejong corpus.

Part-of-speech tagging can be viewed as an intermediate layer.
Starting from surface forms (Column~2), language-specific part-of-speech tagging corresponds to the XPOS annotation in Column~3.
This can be systematically extended to universal part-of-speech tagging by incorporating the \textsc{UPOS} column (Column~4), without reference to constituency bracketing.
In this way, POS tagging remains orthogonal to phrase structure while remaining aligned with both morphological and syntactic representations.

By separating these annotation layers across columns, the proposed format makes explicit which information is required for each task and ensures that constituency parsing and morphological analysis, including part-of-speech tagging, can be performed independently while remaining aligned at the eojeol level.

\begin{figure}[!ht]
    \centering
{
\footnotesize{
\begin{forest}
[S
  [NP-SBJ
[NML    [AdjP [PROPN [프랑스의]]
    [ADJ   [세계적인]]]
    [NML [NOUN [의상]]
    [NOUN [디자이너]]]]
[NP [PROPN [엠마누엘]]
[PROPN [웅가로가]]]]
  [VP
    [NP-AJT
  [NML [NML 
    [NOUN [실내]]
    [NOUN [장식용]]]
    [NOUN [직물]]]
[NOUN [디자이너로]]]
[VERB [나섰다]]
]
[PUNCT [${.}$]]
]
\end{forest}}   
}
\caption{Eojeol-based constituency structure obtained from the Sejong treebank under eojeol-based normalization}
    \label{proposed-sejong}
\end{figure}
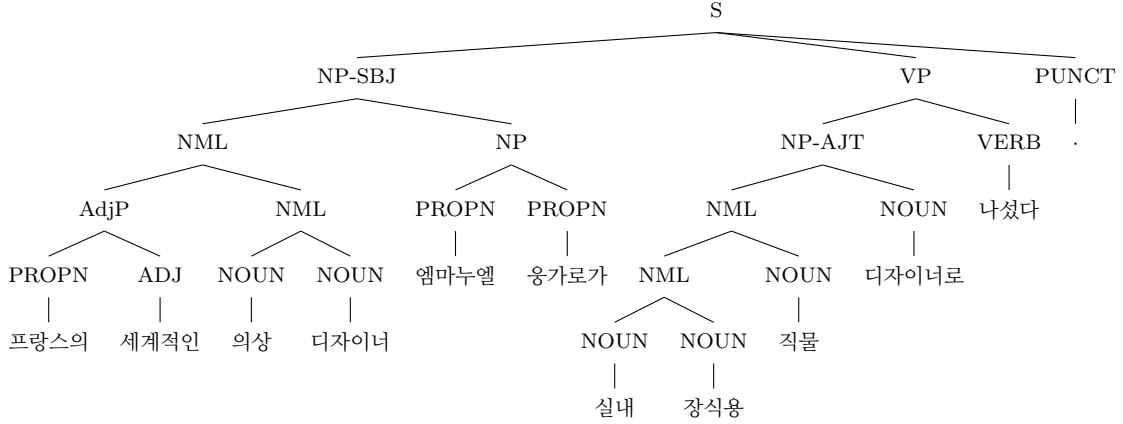

Figure~\ref{proposed-sejong} provides a worked example of the normalization procedure, using the original Sejong tree in Figure~\ref{sejong-treebank-example} as its input.
\label{r2-2-4}
The transformation preserves the overt eojeol terminals and their order, while changing the structural encoding in three respects: recursive binary NP expansions are collapsed, punctuation is represented as an independent eojeol-level terminal, and nominal modifier structure is made explicit through an \texttt{NML} layer.
The example therefore illustrates both the benefit and the cost of normalization.
It yields {a surface-oriented eojeol-based constituency backbone}, but it does not preserve every intermediate branching decision found in the original Sejong tree.

In the subject noun phrase, descriptive material such as 프랑스의  세계적인 의상 디자이너 \textit{peulangseuui} \textit{segyejeogin} \textit{uisang} \textit{dijaineo} (`France-\textsc{gen} world-\textsc{adj} costume designer') is analyzed as a nominal modifier complex encoded within \texttt{NML}, while the proper name 엠마누엘 웅가로가 \textit{emmanuel} \textit{unggaroga} (`Emmanuel Ungaro-\textsc{nom}') forms a separate NP.
This configuration {represents the relation between the descriptive nominal expression and the proper name}, without introducing additional functional projections or altering terminal order.
The refinement preserves the Sejong terminal inventory and linearization, while making the internal modifier domain of the noun phrase explicit and locally interpretable.\footnote{We adopt the analysis shown in Figure~\ref{proposed-sejong} in order to make the internal structure of nominal modification explicit, rather than retaining the following flat structure, which would result from directly debinarizing the original Sejong treebank representation.

\begin{center}
\scriptsize{
\begin{forest}
  [NP-SBJ
    [PROPN [프랑스의]]
    [ADJ [세계적인]]
    [NOUN [의상]]
    [NOUN [디자이너]]
    [PROPN [엠마누엘]]
    [PROPN [웅가로가]]]
\end{forest}}  
\end{center}

Under the proposed analysis in Figure~\ref{proposed-sejong}, 프랑스의 \textit{peulangseuui} modifies 세계적인 \textit{segyejeogin}, and 세계적인 \textit{segyejeogin} in turn modifies 디자이너 \textit{dijaineo}, yielding a stepwise accumulation of adjectival and nominal modifiers within the noun phrase. This configuration reflects one possible analysis of Korean nominal modification, in which modifiers are composed locally and incrementally toward the nominal head, and it aligns with the design principle of encoding nominal modification through an explicit \texttt{NML} layer.

Alternative structural analyses are also linguistically plausible.
In the following figure, 프랑스의 \textit{peulangseuui} continues to modify 세계적인 \textit{segyejeogin}, but 세계적인 \textit{segyejeogin} is analyzed as modifying the referential NP 엠마누엘 웅가로가 \textit{emmanuel unggaroga}, rather than the occupational noun 디자이너 \textit{dijaineo}.
Under this interpretation, 세계적인 \textit{segyejeogin} applies to the individual denoted by the proper name as a whole, instead of contributing to the internal descriptive content of the profession-denoting nominal.
This analysis therefore places adjectival modification at the NP level, outside the local nominal modifier domain encoded by \texttt{NML}.

\begin{center}
\scriptsize{
\begin{forest}
[NP-SBJ
 [AdjP [PROPN [프랑스의]]
    [ADJ   [세계적인]]]
    [NML [NOUN [의상]]
    [NOUN [디자이너]]]
[NP [PROPN [엠마누엘]]
[PROPN [웅가로가]]]]
\end{forest}}  
\end{center}

The following figure illustrates a flatter analysis, in which both 프랑스의 \textit{peulangseuui} and 세계적인 \textit{segyejeogin} attach directly to the referential NP headed by 웅가로가 \textit{unggaroga}.
This configuration is closer to the annotation practice observed in the original Sejong constituency trees and in dependency structures derived from them, where fine-grained internal modifier relations within noun phrases are typically left underspecified and multiple modifiers converge on a single nominal center.

\begin{center}
\scriptsize{
\begin{forest}
[NP-SBJ
 [PROPN [프랑스의]]
    [ADJ   [세계적인]]
    [NML [NOUN [의상]]
    [NOUN [디자이너]]]
[NP [PROPN [엠마누엘]]
[PROPN [웅가로가]]]]
\end{forest}}  
\end{center}

All three analyses are compatible with the surface word order and support coherent semantic interpretations.
In this paper, Figure~\ref{proposed-sejong} is adopted as the representative structure because it renders the internal organization of nominal modification explicit and locally interpretable within the noun phrase.
At the same time, the choice illustrates a broader annotation trade-off: flatter analyses provide greater simplicity and may support higher inter-annotator consistency, whereas \texttt{NML}-based analyses provide greater structural explicitness at the cost of additional attachment decisions.
A systematic empirical comparison, informed by annotation consistency, dependency conversion behavior, and downstream parsing performance, is left for future work.\label{r2-5-3}
}

The resulting tree defines phrase structure over UPOS-labeled eojeol terminals, with morphology represented in a parallel layer and nominal modification encoded locally within noun phrases.

\section{Discussion and implications}

This section discusses the broader implications of the proposed eojeol-based constituency representation.
The discussion focuses on three areas: cross-treebank comparison, alignment between constituency and dependency representations, and future development of Korean linguistic resources.
The section also considers how the proposed representation supports morphology-sensitive analysis while maintaining a stable and interpretable notion of syntactic structure.

\subsection{A small-scale structural validation}
\label{r1-3}

To make the representational consequences of eojeol-based normalization explicit, we include a small-scale structural comparison between original and normalized constituency representations.
The purpose of this comparison is not to evaluate parsing accuracy, but to show how terminal granularity and treebank-specific conventions affect measurable properties of constituency structure.

For each treebank, we compute four descriptive measures over five sample sentences: the number of terminals, the number of nonterminal nodes, terminal depth, and branching factor.
Terminals are counted as morpheme-level leaves in the original morpheme-based representations and as eojeol-level leaves in the normalized representations.
Nonterminals include all internal labeled nodes, including POS or UPOS preterminals.
Depth is averaged over terminals, with the maximum depth reported in parentheses.
Branching factor is averaged over nonterminal nodes, with the maximum branching factor reported in parentheses.
Terminal count reflects the granularity of the yield; nonterminal count reflects the amount of structural material introduced by annotation conventions; depth captures the contribution of binarization, null elements, and morpheme-level projection; and branching factor reflects local tree shape.

\begin{table}[!ht]
\centering
\footnotesize
\begin{tabular}{lrrrr}
\toprule
& Avg. terminals  & Avg. nonterminals & Avg. depth & Avg. branching  \\
 & (total) & (total) &  (max) & (max) \\
\midrule

Original Sejong & 18.20 (91) & 34.40 (172) & 5.47 (8) & 1.50 (4)  \\
Normalized Sejong & 10.20 (51) & 18.80 (94) & 4.00 (7) & 1.49 (6)  \\\midrule

Original Penn Korean & 30.20 (151) & 50.80 (254) & 6.89 (12) & 1.57 (6)  \\
Normalized Penn Korean & 15.20 (76) & 31.60 (158) & 6.42 (11) & 1.45 (5)  \\\midrule

Original KAIST & 18.40 (92) & 32.80 (164) & 5.65 (11) & 1.53 (4)  \\
Normalized KAIST & 10.20 (51) & 23.20 (116) & 5.12 (8) & 1.40 (4)  \\\bottomrule

\end{tabular}
\caption{Structural comparison before and after eojeol-based normalization}
\label{tab:structural-validation}
\end{table}

Table~\ref{tab:structural-validation} shows that eojeol-based normalization primarily affects the yield and the amount of projected structural material. Across all three resources, terminal counts decrease because morpheme-level leaves, null elements, or separately projected morphological units are collapsed into overt eojeol-level terminals. Nonterminal counts also decrease, but for different resource-specific reasons: debinarization in Sejong, null-element removal and terminal realignment in Penn Korean, and the collapse of morpheme-level projections in KAIST. These reductions should therefore be read not as improvements in tree quality, but as the structural consequences of placing different resources over a common terminal domain.

Depth and branching factor show a more differentiated pattern. The strongest depth reduction appears where normalization removes annotation-induced structure, especially binarization or morpheme-level projection. By contrast, average branching factor changes only modestly, although local maxima may increase when debinarization creates flatter configurations. This indicates that normalization does not erase the phrase-structural organization of each resource. Rather, it removes resource-specific representational material that intervenes between the surface string and the constituency backbone.

The expected effect of normalization is therefore not that the three treebanks become structurally identical, but that they become comparable over a shared eojeol-based yield. Debinarizing Sejong reduces depth introduced by Chomsky-normal-form annotation. Removing null elements from Penn Korean eliminates abstract terminals that do not correspond to overt eojeol. Collapsing morpheme-level terminals in KAIST reduces the mismatch between morphological segmentation and constituency yield. Across these transformations, the normalized representations preserve the surface eojeol sequence and phrase-level constituency relations, while abstracting away from resource-specific choices concerning binarity, null elements, and the syntactic projection of functional morphology.

This small-scale comparison gives operational support to the proposed resource design. Eojeol-based normalization produces measurable changes in structural complexity and terminal alignment, but it should not be interpreted as a ranking of one treebank over another. Once the original representations are normalized to eojeol-level terminals, the constituency yield can be directly aligned with eojeol-based dependency tokens without reconstructing morpheme-level boundaries. The validation therefore does not test parsing accuracy, but verifies a necessary precondition for cross-treebank comparison and constituency–dependency alignment: a shared, stable terminal sequence.
\label{r2-3}

\subsection{Implications for cross-treebank comparison}

An eojeol-based constituency representation provides a common structural basis for comparing Korean constituency treebanks that differ substantially in annotation practice.
By abstracting away from binarization constraints, null elements, and morpheme-level terminals, the proposed normalization yields constituency structures defined over the same terminal units.
This makes it possible to compare phrase structure across Sejong, Penn Korean, and KAIST-derived resources without reconstructing terminals or redefining syntactic relations.

Such comparability is particularly important for empirical analysis.
Differences observed across treebanks can be attributed more directly to representational choices, such as labeling conventions or attachment decisions, rather than to artifacts introduced by binarization or tokenization.
As a result, cross-resource evaluation and qualitative comparison become more interpretable, and aggregate analyses over multiple treebanks become feasible with fewer resource-specific preprocessing steps.

\subsection{Implications for constituency--dependency alignment}

The proposed representation also clarifies the interface between constituency and dependency representations.
Because eojeol are the basic terminals of the proposed constituency structure and of widely used Korean dependency resources, defining constituency over eojeol establishes a direct correspondence between the two formalisms.
Phrase structure relations can therefore be related to dependency arcs without reconstructing word boundaries or collapsing morpheme-level nodes.

This alignment can support conversion between constituency and dependency representations, but conversion is not the primary claim of the proposal.
Rather, the central point is that both representations can be interpreted over the same surface-form terminal domain.
Constituency-to-dependency conversion can proceed by interpreting phrase structure over eojeol terminals, while dependency-to-constituency conversion can project hierarchical groupings without introducing morpheme-level terminals or auxiliary projections.
The result is a more transparent mapping between hierarchical and relational representations, grounded in a shared notion of terminal unit.

More generally, separating constituency from morphology reduces the risk that alignment or conversion procedures inadvertently encode morphological assumptions as syntactic structure.
Dependency relations remain defined over eojeol, and morphological information can be consulted when needed, but it does not determine the shape of the constituency representation.

\subsection{Implications for future Korean resource development and availability}

The proposed eojeol-based constituency scheme provides a principled framework for future Korean resource development by explicitly separating syntactic structure from morphological analysis.
Constituency treebanks can be constructed or extended using eojeol-level terminals and UPOS-labeled preterminals, while morphological segmentation and fine-grained part-of-speech information are maintained in a parallel representational layer.
This separation avoids conflating morphology with phrase structure and clarifies the role of each level of annotation within the overall resource architecture.

Such a design supports incremental and modular corpus development.
Morphological annotation schemes can be revised, enriched, or diversified without requiring corresponding changes to constituency structure, and alternative morphological analyses can coexist over the same syntactic backbone.
Conversely, syntactic annotation can be refined or extended independently of decisions about morphological segmentation.
This modularity facilitates long-term maintenance and reuse of Korean linguistic resources, particularly in settings where annotation standards or analytical goals evolve over time.

At the same time, the explicit availability of aligned morphological information enables morphology-sensitive downstream applications.
Tasks such as generation, grammatical error analysis, and morphosyntactic disambiguation can exploit detailed morphological annotation, while the constituency representation remains structurally stable and comparable across resources.
In this respect, the proposed scheme balances linguistic expressiveness with representational clarity, offering a foundation for systematic corpus construction and comparative analysis.

The practical scope of this proposal is nonetheless shaped by the licensing conditions of existing Korean constituency treebanks.
Major resources, including Sejong, Penn Korean, and KAIST, are subject to copyright restrictions that limit redistribution and large-scale public release.
As a result, openly available constituency data directly conforming to the proposed representation remain limited.
This constraint highlights the importance of alternative pathways for resource expansion, particularly the conversion of publicly available dependency corpora into eojeol-based constituency representations.
Such conversions would make it possible to instantiate the proposed framework in openly distributable resources, mitigating licensing barriers while preserving compatibility with established Korean syntactic annotations.

\section{Conclusion}

This paper has argued for an eojeol-based terminal representation in Korean constituency annotation, with morpheme-level information encoded in a separate morphological layer.
This representational separation preserves an interpretable notion of phrase structure, facilitates comparison and alignment across existing Korean treebanks, and provides a shared interface with eojeol-based dependency resources.
More broadly, the paper develops a resource-design framework for Korean constituency treebank construction that separates syntactic representation from morphological annotation while maintaining compatibility with existing resources.
At the same time, rich morphological annotation can be retained without redefining constituency terminals, providing a stable foundation for future Korean resource development.

\section*{Acknowledgments}
This work was supported by the Institute of Information \& Communications Technology Planning \& Evaluation (IITP) grant funded by the Korea government (MSIT) (No. RS-2025-25441313, Professional AI Talent Development Program for Multimodal AI Agents).


\begin{thebibliography}{29}
\providecommand{\natexlab}[1]{#1}
\providecommand{\url}[1]{\texttt{#1}}
\expandafter\ifx\csname urlstyle\endcsname\relax
  \providecommand{\doi}[1]{doi: #1}\else
  \providecommand{\doi}{doi: \begingroup \urlstyle{rm}\Url}\fi

\bibitem[Abeill{\'{e}} et~al.(2003)Abeill{\'{e}}, Cl{\'{e}}ment, and Toussenel]{abeille-clement-toussenel:2003}
Anne Abeill{\'{e}}, Lionel Cl{\'{e}}ment, and François Toussenel.
\newblock {Building a Treebank for French}.
\newblock In Anne Abeill{\'{e}}, editor, \emph{Treebanks: Building and Using Parsed Corpora}, pages 165--188. Kluwer, 2003.

\bibitem[Bikel(2004)]{bikel:2004:CL}
Daniel~M. Bikel.
\newblock {Intricacies of Collins' Parsing Model}.
\newblock \emph{Computational Linguistics}, 30\penalty0 (4):\penalty0 479--511, 2004.
\newblock \doi{10.1162/0891201042544929}.
\newblock URL \url{https://doi.org/10.1162/0891201042544929}.

\bibitem[Choi et~al.(2012)Choi, Park, and Choi]{choi-park-choi:2012:SP-SEM-MRL}
DongHyun Choi, Jungyeul Park, and Key-Sun Choi.
\newblock {Korean Treebank Transformation for Parser Training}.
\newblock In \emph{Proceedings of the ACL 2012 Joint Workshop on Statistical Parsing and Semantic Processing of Morphologically Rich Languages}, pages 78--88, Jeju, Republic of Korea, 2012. Association for Computational Linguistics.
\newblock URL \url{http://www.aclweb.org/anthology/W12-3411}.

\bibitem[Choi et~al.(1994)Choi, Han, Han, and Kwon]{choi-EtAl:1994}
Key-Sun Choi, Young~S. Han, Young~G. Han, and Oh~W. Kwon.
\newblock {KAIST Tree Bank Project for Korean: Present and Future Development}.
\newblock In \emph{Proceedings of the International Workshop on Sharable Natural Language Resources}, pages 7--14, Nara Institute of Science and Technology, 1994. Nara Institute of Science and Technology.

\bibitem[Chung et~al.(2010)Chung, Post, and Gildea]{chung-post-gildea:2010:SPMRL}
Tagyoung Chung, Matt Post, and Daniel Gildea.
\newblock {Factors Affecting the Accuracy of Korean Parsing}.
\newblock In \emph{Proceedings of the NAACL HLT 2010 First Workshop on Statistical Parsing of Morphologically-Rich Languages}, pages 49--57, Los Angeles, CA, USA, 2010. Association for Computational Linguistics.
\newblock URL \url{http://www.aclweb.org/anthology/W10-1406}.

\bibitem[Cocke(1969)]{cocke:1969}
John Cocke.
\newblock \emph{{Programming Languages and Their Compilers: Preliminary Notes}}.
\newblock New York University, USA, 1969.
\newblock ISBN B0007F4UOA.

\bibitem[de~Marneffe et~al.(2021)de~Marneffe, Manning, Nivre, and Zeman]{de-marneffe-etal-2021-universal}
Marie-Catherine de~Marneffe, Christopher~D. Manning, Joakim Nivre, and Daniel Zeman.
\newblock {Universal Dependencies}.
\newblock \emph{Computational Linguistics}, 47\penalty0 (2):\penalty0 255--308, 6 2021.
\newblock \doi{10.1162/coli{\_}a{\_}00402}.
\newblock URL \url{https://aclanthology.org/2021.cl-2.11}.

\bibitem[Han et~al.(2002)Han, Han, Ko, Palmer, and Yi]{han-EtAl:2002}
Chung-Hye Han, Na-Rae Han, Eon-Suk Ko, Martha Palmer, and Heejong Yi.
\newblock {Penn Korean Treebank: Development and Evaluation}.
\newblock In \emph{Proceedings of the 16th Pacific Asia Conference on Language, Information and Computation}, pages 69--78, Jeju, Korea, 2002. Pacific Asia Conference on Language, Information and Computation.

\bibitem[Han et~al.(2020{\natexlab{a}})Han, Kim, Moulton, and Lidz]{han-etal-2020-null}
Chung-hye Han, Kyeong-min Kim, Keir Moulton, and Jeffrey Lidz.
\newblock {Null Objects in Korean: Experimental Evidence for the Argument Ellipsis Analysis}.
\newblock \emph{Linguistic Inquiry}, 51\penalty0 (2):\penalty0 319--340, 2020{\natexlab{a}}.
\newblock ISSN 0024-3892.
\newblock \doi{10.1162/ling{\_}a{\_}00342}.
\newblock URL \url{https://doi.org/10.1162/ling%5C_a%5C_00342}.

\bibitem[Han et~al.(2020{\natexlab{b}})Han, Oh, Jin, and Kim]{han-etal-2020-annotation}
Ji~Yoon Han, Tae~Hwan Oh, Lee Jin, and Hansaem Kim.
\newblock {Annotation Issues in Universal Dependencies for Korean and Japanese}.
\newblock In Marie-Catherine de~Marneffe, Miryam de~Lhoneux, Joakim Nivre, and Sebastian Schuster, editors, \emph{Proceedings of the Fourth Workshop on Universal Dependencies (UDW 2020)}, pages 99--108, Barcelona, Spain (Online), 12 2020{\natexlab{b}}. Association for Computational Linguistics.
\newblock URL \url{https://aclanthology.org/2020.udw-1.12}.

\bibitem[Kasami(1966)]{kasami:1966}
Tadao Kasami.
\newblock {An Efficient Recognition and Syntax-Analysis Algorithm for Context-Free Languages}.
\newblock Technical report, University of Illinois at Urbana-Champaign, 3 1966.
\newblock URL \url{http://hdl.handle.net/2142/74304}.

\bibitem[Kim et~al.(2024)Kim, Chen, Jo, Lim, Park, and Park]{kim-etal-2024-kud}
Kyuwon Kim, Yige Chen, Eunkyul~Leah Jo, KyungTae Lim, Jungyeul Park, and Chulwoo Park.
\newblock {K-UD: Revising Korean Universal Dependencies Guidelines}.
\newblock \emph{arXiv}, pages 1--6, 2024.
\newblock URL \url{https://arxiv.org/abs/2412.00856}.

\bibitem[Kim and Park(2022)]{kim-park:2022}
Mija Kim and Jungyeul Park.
\newblock {A note on constituent parsing for Korean}.
\newblock \emph{Natural Language Engineering}, 28\penalty0 (2):\penalty0 199--222, 2022.
\newblock \doi{10.1017/S1351324920000479}.

\bibitem[Marcus et~al.(1993)Marcus, Marcinkiewicz, and Santorini]{marcus-etal-1993-building}
Mitchell~P. Marcus, Mary~Ann Marcinkiewicz, and Beatrice Santorini.
\newblock {Building a Large Annotated Corpus of English: The Penn Treebank}.
\newblock \emph{Computational linguistics}, 19\penalty0 (2):\penalty0 313--330, 1993.
\newblock URL \url{https://aclanthology.org/J93-2004}.

\bibitem[McDonald et~al.(2013)McDonald, Nivre, Quirmbach-Brundage, Goldberg, Das, Ganchev, Hall, Petrov, Zhang, T{\"{a}}ckstr{\"{o}}m, Bedini, Bertomeu~Castell{\'{o}}, and Lee]{mcdonald-etal-2013-universal}
Ryan McDonald, Joakim Nivre, Yvonne Quirmbach-Brundage, Yoav Goldberg, Dipanjan Das, Kuzman Ganchev, Keith Hall, Slav Petrov, Hao Zhang, Oscar T{\"{a}}ckstr{\"{o}}m, Claudia Bedini, Núria Bertomeu~Castell{\'{o}}, and Jungmee Lee.
\newblock {Universal Dependency Annotation for Multilingual Parsing}.
\newblock In Hinrich Schuetze, Pascale Fung, and Massimo Poesio, editors, \emph{Proceedings of the 51st Annual Meeting of the Association for Computational Linguistics (Volume 2: Short Papers)}, pages 92--97, Sofia, Bulgaria, 8 2013. Association for Computational Linguistics.
\newblock URL \url{https://aclanthology.org/P13-2017}.

\bibitem[Noh et~al.(2018)Noh, Han, Oh, and Kim]{noh-etal-2018-enhancing}
Youngbin Noh, Jiyoon Han, Tae~Hwan Oh, and Hansaem Kim.
\newblock {Enhancing Universal Dependencies for Korean}.
\newblock In \emph{Proceedings of the Second Workshop on Universal Dependencies (UDW 2018)}, pages 108--116, Brussels, Belgium, 11 2018. Association for Computational Linguistics.
\newblock \doi{10.18653/v1/W18-6013}.
\newblock URL \url{https://aclanthology.org/W18-6013}.

\bibitem[Park(2018)]{chulwoo-2019-for-developing}
Chulwoo Park.
\newblock {For Developing the Synthetic Perspective about the Concept of Word in Korean}.
\newblock \emph{HANGEUL}, 79\penalty0 (2):\penalty0 327--368, 2018.
\newblock ISSN 1225-0449.
\newblock \doi{https://doi.org/10.22557/HG.2018.06.79.2.327}.
\newblock URL \url{https://www.kci.go.kr/kciportal/ci/sereArticleSearch/ ciSereArtiView.kci? sereArticleSearchBean.artiId=ART002356650}.

\bibitem[Park and Kim(2023)]{park-kim-2023-role}
Jungyeul Park and Mija Kim.
\newblock {A role of functional morphemes in Korean categorial grammars}.
\newblock \emph{Korean Linguistics}, 19\penalty0 (1):\penalty0 1--30, 2023.
\newblock \doi{10.1075/kl.22003.par}.
\newblock URL \url{https://doi.org/10.1075/kl.22003.par}.

\bibitem[Park and Kim(2024)]{park-kim-2024-word}
Jungyeul Park and Mija Kim.
\newblock {Word segmentation granularity in Korean}.
\newblock \emph{Korean Linguistics}, 20\penalty0 (1):\penalty0 83--113, 2024.
\newblock URL \url{https://benjamins.com/catalog/kl.00008.par}.

\bibitem[Park and Tyers(2019)]{park-tyers:2019:LAW}
Jungyeul Park and Francis Tyers.
\newblock {A New Annotation Scheme for the Sejong Part-of-speech Tagged Corpus}.
\newblock In \emph{Proceedings of the 13th Linguistic Annotation Workshop}, pages 195--202, Florence, Italy, 8 2019. Association for Computational Linguistics.
\newblock URL \url{https://www.aclweb.org/anthology/W19-4022}.

\bibitem[Park et~al.(2016)Park, Hong, and Cha]{park-hong-cha:2016:PACLIC}
Jungyeul Park, Jeen-Pyo Hong, and Jeong-Won Cha.
\newblock {Korean Language Resources for Everyone}.
\newblock In \emph{Proceedings of the 30th Pacific Asia Conference on Language, Information and Computation: Oral Papers (PACLIC 30)}, pages 49--58, Seoul, Korea, 2016. Pacific Asia Conference on Language, Information and Computation.
\newblock URL \url{http://aclweb.org/anthology/Y/Y16/Y16-2002.pdf}.

\bibitem[Park et~al.(2021)Park, Moon, Kim, Cho, Han, Park, Song, Kim, Song, Oh, Lee, Oh, Lyu, Jeong, Lee, Seo, Lee, Kim, Lee, Jang, Do, Kim, Lim, Lee, Park, Shin, Kim, Park, Oh, Ha, and Cho]{park-etal-2021-klue}
Sungjoon Park, Jihyung Moon, Sungdong Kim, Won~Ik Cho, Ji~Yoon Han, Jangwon Park, Chisung Song, Junseong Kim, Youngsook Song, Taehwan Oh, Joohong Lee, Juhyun Oh, Sungwon Lyu, Younghoon Jeong, Inkwon Lee, Sangwoo Seo, Dongjun Lee, Hyunwoo Kim, Myeonghwa Lee, Seongbo Jang, Seungwon Do, Sunkyoung Kim, Kyungtae Lim, Jongwon Lee, Kyumin Park, Jamin Shin, Seonghyun Kim, Lucy Park, Alice Oh, Jung-Woo Ha, and Kyunghyun Cho.
\newblock {KLUE: Korean Language Understanding Evaluation}.
\newblock In Joaquin Vanschoren and Serena Yeung, editors, \emph{Proceedings of the Neural Information Processing Systems Track on Datasets and Benchmarks}, volume~1, pages 1--25. Curran, 2021.
\newblock URL \url{https://datasets-benchmarks-proceedings.neurips.cc/ paper_files/paper/2021/file/98dce83da57b0395e163467c9dae521b- Paper-round2.pdf}.

\bibitem[Petrov et~al.(2012)Petrov, Das, and McDonald]{petrov-das-mcdonald:2012:LREC}
Slav Petrov, Dipanjan Das, and Ryan McDonald.
\newblock {A Universal Part-of-Speech Tagset}.
\newblock In \emph{Proceedings of the Eighth International Conference on Language Resources and Evaluation (LREC-2012)}, pages 2089--2096, Istanbul, Turkey, 2012. European Language Resources Association (ELRA).
\newblock ISBN 978-2-9517408-7-7.

\bibitem[Seddah et~al.(2013)Seddah, Tsarfaty, K{\"{u}}bler, Candito, Choi, Farkas, Foster, Goenaga, Gojenola~Galletebeitia, Goldberg, Green, Habash, Kuhlmann, Maier, Nivre, Przepi{\'{o}}rkowski, Roth, Seeker, Versley, Vincze, Woli{\'{n}}ski, Wr{\'{o}}blewska, and de~la Clergerie]{seddah-EtAl:2013:SPMRL}
Djamé Seddah, Reut Tsarfaty, Sandra K{\"{u}}bler, Marie Candito, Jinho~D. Choi, Richárd Farkas, Jennifer Foster, Iakes Goenaga, Koldo Gojenola~Galletebeitia, Yoav Goldberg, Spence Green, Nizar Habash, Marco Kuhlmann, Wolfgang Maier, Joakim Nivre, Adam Przepi{\'{o}}rkowski, Ryan Roth, Wolfgang Seeker, Yannick Versley, Veronika Vincze, Marcin Woli{\'{n}}ski, Alina Wr{\'{o}}blewska, and Eric~Villemonte de~la Clergerie.
\newblock {Overview of the SPMRL 2013 Shared Task: A Cross-Framework Evaluation of Parsing Morphologically Rich Languages}.
\newblock In \emph{Proceedings of the Fourth Workshop on Statistical Parsing of Morphologically-Rich Languages}, pages 146--182, Seattle, Washington, USA, 10 2013. Association for Computational Linguistics.
\newblock URL \url{http://www.aclweb.org/anthology/W13-4917}.

\bibitem[Seo et~al.(2019)Seo, Kim, Sung, and Yoo]{seo-etal-2019-ud}
Saetbyol Seo, Myeong-ju Kim, YeonSook Sung, and Seong~Hee Yoo.
\newblock {A Proposal on Universal Dependencies (v.2) Annotation for Korean}.
\newblock \emph{Language and Information}, 23\penalty0 (1):\penalty0 91--122, 2019.
\newblock URL \url{https://doi.org/10.29403/LI.23.1.5}.

\bibitem[Vadas and Curran(2011)]{vadas-curran-2011-parsing}
David Vadas and James~R Curran.
\newblock {Parsing Noun Phrases in the Penn Treebank}.
\newblock \emph{Computational Linguistics}, 37\penalty0 (4):\penalty0 753--809, 12 2011.
\newblock \doi{10.1162/COLI{\_}a{\_}00076}.
\newblock URL \url{https://aclanthology.org/J11-4006/}.

\bibitem[Xia and Palmer(2001)]{xia-palmer-2001-converting}
Fei Xia and Martha Palmer.
\newblock {Converting Dependency Structures to Phrase Structures}.
\newblock In \emph{Proceedings of the First International Conference on Human Language Technology Research}, San Diego, CA, 2001. Association for Computational Linguistics.
\newblock URL \url{http://www.aclweb.org/anthology/H01-1014}.

\bibitem[Xue et~al.(2005)Xue, Xia, Chiou, and Palmer]{xue-etal-2005-ctb}
Nianwen Xue, Fei Xia, Fu-dong Chiou, and Marta Palmer.
\newblock {The Penn Chinese TreeBank: Phrase Structure Annotation of a Large Corpus}.
\newblock \emph{Natural Language Engineering}, 11\penalty0 (2):\penalty0 207--238, 6 2005.
\newblock ISSN 1351-3249.
\newblock \doi{10.1017/S135132490400364X}.
\newblock URL \url{https://doi.org/10.1017/S135132490400364X}.

\bibitem[Younger(1967)]{younger:1967}
Daniel~H. Younger.
\newblock {Recognition and parsing of context-free languages in time n3}.
\newblock \emph{Information and Control}, 10\penalty0 (2):\penalty0 189--208, 1967.
\newblock ISSN 0019-9958.
\newblock \doi{https://doi.org/10.1016/S0019-9958(67)80007-X}.
\newblock URL \url{http://www.sciencedirect.com/science/article/pii/ S001999586780007X}.

\end{thebibliography}

\end{document}